\documentclass[default,iicol]{sn-jnl}
\usepackage[T1]{fontenc}

\jyear{2023}%

\theoremstyle{thmstyleone}%
\theoremstyle{thmstyletwo}%
\newtheorem{example}{Example}%

\theoremstyle{thmstylethree}%
\newtheorem{definition}{Definition}%

\usepackage{tabularx}
\usepackage{makecell}
\usepackage[T1]{fontenc}

\DeclareUnicodeCharacter{2113}{$\ell$}

\usepackage{adjustbox}
\definecolor{myblue}{HTML}{4472c4}
\newenvironment{colbox}{%
    \noindent
    \adjustbox{innerenv={varwidth}[c]{0.92\columnwidth},margin=\fboxsep+.03\columnwidth \fboxsep+.03\columnwidth,bgcolor=myblue,center}\bgroup \color{white}
}{%
    \egroup
}

\newcommand{\oq}[1]{{\footnotesize\fcolorbox{magenta}{white}{\textcolor{magenta}{\textbf{#1}}}}}

\newcommand{\takeaway}[1]{\leavevmode\newline\begin{colbox}\textbf{Takeaway:} #1 \end{colbox}}

\begin{document}
\author{Boris van Breugel, Mihaela van der Schaar}
\title[\ ]{Beyond Privacy: Navigating the Opportunities and Challenges of Synthetic Data}

\abstract{
Generating synthetic data through generative models is gaining interest in the ML community and beyond. In the past, synthetic data was often regarded as a means to private data release, but a surge of recent papers explore how its potential reaches much further than this---from creating more fair data to data augmentation, and from simulation to text generated by ChatGPT. In this perspective we explore whether, and how, synthetic data may become a dominant force in the machine learning world, promising a future where datasets can be tailored to individual needs. Just as importantly, we discuss which fundamental challenges the community needs to overcome for wider relevance and application of synthetic data---the most important of which is quantifying how much we can trust any finding or prediction drawn from synthetic data.
}

\maketitle

\section{Introduction} \label{sec:intro}
\textbf{Motivation.}
Data is the foundation of most science, but real data can be severely limiting. It may be privacy-sensitive, unfair, unbalanced, unrepresentative, or it may simply not exist. In turn, these limitations continuously constrain how the ML community operates---from training to testing, from model development to deployment. Do we need to accept this as reality, or is there another way? The answer might lie in synthetic data.

\begin{figure}[hbt]
    \centering
    \includegraphics[width=\columnwidth]{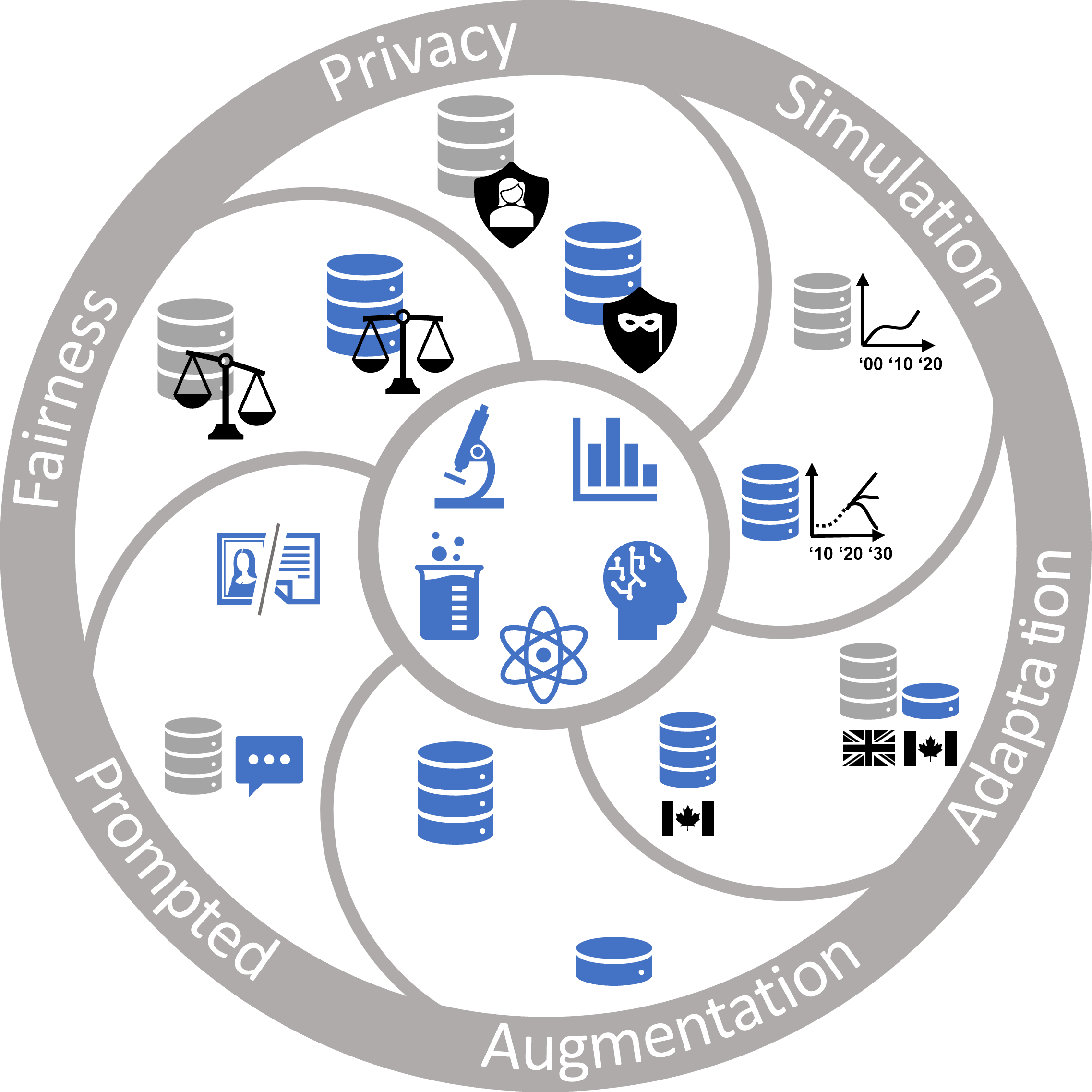}
    \caption{Overview synthetic data use cases}
    \label{fig:use_cases}
\end{figure}

Advances in deep generative modelling have seen a steep rise in methods that aim to replace real data with synthetic data. The first uses of synthetic data were mostly privacy-focused, aiming to create realistic synthetic data that mimics the real data but does not disclose sensitive information \cite{Ho2021DP-GAN:Nets, Yoon2020AnonymizationADS-GAN,Jordon2019PATE-GAN:Guarantees}. More recently, however, there has been an increasing interest in extending synthetic data to use cases where it \emph{improves} upon real data (see Fig. \ref{fig:use_cases}), for example providing better fairness \cite{xu2018fairgan,xu2019achieving,vanBreugel2021DECAF:Networks}, augmenting the dataset size \citep{Antoniou2017DataNetworks, dina2022effect, das2022conditional, Bing2022ConditionalPopulations}, and creating or simulating data for different domains \cite{Yoon2018RadialGAN:Networks, wang2019LearningWild}. The latest and most widely acclaimed development is user-prompted data (e.g. OpenAI's DALL-E and ChatGPT), of which the use cases and influences on a wider audience are hard to understate.

In this perspective, we reflect on these advances, and discuss opportunities and challenges that lie ahead. We hope to shed light on whether the ML community requires a radical paradigm shift: away from the uncountable, inherent limitations of real data, towards an ML world where realistic, customizable, synthetic data becomes the lead actor. We will find that though this future looks bright, there are fundamental challenges of synthetic data that need to be urgently overcome before synthetic data can be trusted and used more widely.

\textbf{Defining synthetic data.} We focus on data-driven synthetic data, which we define as follows:
\begin{definition}
    Given a stochastic mathematical model that outputs data and is fitted on real data with the purpose of describing and mimicking (some part of) the real data's distribution. The data generated by such a model we call \textbf{data-driven synthetic data}.
\end{definition}

\begin{table*}[hbt]
    \centering
    \small
    \caption{Use cases of synthetic data}
    \label{tab:use_cases}
    \begin{tabularx}{\textwidth}{l|clX} \toprule
         Use case & Data input & Other input & Aim  \\ \midrule
         Privacy & $\mathcal{D}_{T}\sim p_T$ & \makecell[tl]{Privacy\\ constraints \\ (e.g. DP)} & Generate synthetic data that follows the same real data distribution, but does not disclose information \\ \hline
         Augmentation & $\mathcal{D}_{T}\sim p_T$ & - & Generate more data from $X\sim p_T$, possibly to balance highly underrepresented classes or groups.\\ \hline
         Adaptation & \makecell[tc]{$\mathcal{D}_{S}\sim p_S$, \\$\mathcal{D}_{T}\sim p_T$,\\ $ \vert \mathcal{D}_{S} \vert  \gg  \vert\mathcal{D}_T\vert $} & - & Generate data from target distribution $p_T$, using mostly data from other domain(s) $p_S$. \\ \hline
         Simulation & $\mathcal{D}_{S}\sim p_S$ & \makecell[tl]{Shift \\ parameters} & Generate data from simulated domain, using data from original source domain $\mathcal{D}_S$ and possible shift or simulation input. \\ \hline
         Fairness & $\mathcal{D}_{S}\sim p_S$ & \makecell[tl]{Fairness \\ notion} & Generate fair synthetic data, using unfair real data and fairness constraints. \\ \hline
         User-prompted & $\mathcal{D}_{T}\sim p_T$ & Prompt & Generate samples conditionally on a prompt (e.g. ChatGPT, DALL-E).\\
     \bottomrule
    \end{tabularx}
\end{table*}

We use the \textit{data-driven} constraint to focus ourselves on recent ML deep generative models, and \textbf{not} traditional, \emph{hand-crafted} synthetic data---datasets generated by user-defined mathematical equations or physics simulations that are often over-simplified. The ``(some part of)'' also implies we are often not only interested in mimicking the real data---additional input enables the generation of synthetic data with very different characteristics than the real data, allowing the creation of purpose-built datasets. For example, by modelling the data's distribution conditional on age, we can generate data for an older or younger population than the original. Henceforth, we will drop ``data-driven'' and simply refer to this as synthetic data. 

\textbf{Focus.} 
In the perspective we do not discuss specific generative model architectures (e.g. VAEs \citep{Rezende2014StochasticModels,Kingma2014Auto-EncodingBayes}, GANs \citep{Goodfellow2014GenerativeNetworks}, score-based/diffusion models \citep{Sohl-Dickstein2015DeepThermodynamics,Song2019GenerativeDistribution, Ho2020DenoisingModels}), for which good reviews exist (e.g. see \citep{Bond-Taylor2021DeepModels}). Similarly, although many synthetic data works focus primarily on computer vision, we will not make a distinction between different data modalities---we will see that many of the key applications or issues are modality-independent.

The perspective is split in two parts. In Section \ref{sec:use_cases} we discuss the main use cases of synthetic data with specific opportunities and challenges for each. Subsequently, in Section \ref{sec:challenges} we highlight general challenges and opportunities.

\section{Use cases}  \label{sec:use_cases}
This section is structured by the main use cases of synthetic data, see Figure \ref{fig:use_cases} and Table \ref{tab:use_cases}. In Section \ref{sec:privacy} we discuss the most traditional use case:  privacy. Continuing to a more recent strand of research, we discuss three uses of synthetic for settings where real data is insufficient or non-existent. We start at \textit{augmentation} (Section \ref{sec:augmentation}), where we have some data for the target domain of interest $T$, but would like to create more data for this domain in a generative manner. Subsequently, we consider \textit{domain adaptation} (Section \ref{sec:adaptation}), where we again assume we have some data for domain $T$, but we also have data from some related domain $S$ that we can leverage. Continuing, we introduce \textit{data-driven simulation} (Section \ref{sec:simulation}), where we assume we have data for some domain $S$, and we use prior knowledge or assumptions on possible distributional shifts to generate data for a different domain, $T$. 

Furthermore, we explore synthetic data for \textit{fairness} (Section \ref{sec:fairness}), where the aim is to debias data. At last, in Section \ref{sec:prompted} we discuss user-prompted synthetic data (e.g. ChatGPT). 

\subsection{Privacy} \label{sec:privacy}
\textbf{Motivation.} Many real-world datasets contain private information: personal, sensitive data about individuals. Sharing this data may not be possible due to the risk of violating data privacy, which in turn impedes scientific research, reproducibility, and ML development itself. Synthetic data is a potential solution. It aims to generate data that has the same distribution as the original data, but that does not disclose information about individuals. This signifies a stark contrast with traditional anonymization methods, which often risk either not being fully private, e.g. allowing linkage attacks \cite{Rigaki2020ALearning}, or losing too much utility (e.g. when features are coarsened or made noisy). 

Synthetic data is not private by default; a generative model could overfit to its training data---learning to memorise samples or disclose sensitive statistics. As a result, a large body of work has focused on how one can adapt generative models to ensure privacy. Overall, the idea is usually to limit how much any sample can influence the final dataset (e.g. \citep{Dwork2006OurGeneration, Ho2021DP-GAN:Nets}, or to avoid that synthetic data is highly overfitted to real data (e.g. \citep{Yoon2020AnonymizationADS-GAN}). It is beyond the scope of this article to review these works, instead we refer to Appendix \ref{app:privacy} for a high-level overview and \cite{Stadler2022SyntheticDay} for a more detailed discussion. 

\textbf{Challenges.} 
Many challenges remain for private synthetic data. First, there is no perfect metric or definition for measuring or guaranteeing privacy. Typical anonymization metrics (e.g. k-anonymity) rely on linkage,\footnote{Linkage attacks focus on the ability to identify someone in an anonymized dataset by using some additional knowledge. For example, assume we have access to anonymized hospital records that contain some person we are interested in and we know this person is 1m65 tall, weighs 70kg, and lives in some postcode; we can go through the dataset and might be able to find this person's record and acquire sensitive medical information.} but linkage is not an issue for generative models since data is created from scratch. Similarly, though differential privacy \citep{Dwork2006OurGeneration} is a popular definition for privacy, it has serious disadvantages; it is non-intuitive, cannot be measured directly and it is unclear whether it presents a good trade-off of privacy and utility. Attacker models can be used in a white-hat-hacker-fashion for quantifying privacy vulnerability, but since these methods inherently rely on some other ML attacker model (including its parameterisation and training process), more research is required into these attackers' reliability, robustness, and benchmarking capabilities---essential properties for a good metric. We discuss metrics further in Section \ref{sec:challenges}.

Secondly, there is a privacy-utility trade-off: generating synthetic data with privacy constraints often leads to slightly noisier distributions, leading to a loss of data utility. Finding the right balance is hard \cite{Stadler2022SyntheticDay}, not least because both privacy and utility are hard to quantify. Developing better and more interpretable metrics will be an essential first step for navigating this trade-off.

Thirdly, it is hard to guarantee privacy in a future-proof fashion; something that seems private now, may change in the future when attacker models improve and attackers may have access to more data (e.g. as showcased in \cite{vanBreugel2023domias}). 

\takeaway{Synthetic data has a potential to replace privacy-sensitive real data, thereby allowing less red tape, better reproducibility, and more accessible science. On the other hand, there exists a privacy-utility trade-off that is hard to navigate. }

\subsection{Augmentation and balancing} \label{sec:augmentation}
\textbf{Motivation.}
Data scarcity can be a significant challenge in the real-world, which effectively limits how strong an ML model we can train without overfitting. Synthetic data can be used to mitigate this, by artificially increasing the size of real datasets. 

Augmentation is a popular technique in ML that often leads to better downstream models. In the image and text domain, one usually uses prior knowledge about invariances to augment training data; e.g. rotate images under the assumption that a rotated object is still the same object. Unfortunately, such prior knowledge is harder to acquire in other domains, for example for tabular data. Instead, one can use a generative model for augmentation. The advantage over traditional tabular data augmentation methods (e.g. SMOTE \citep{Chawla2002SMOTE:Technique}) is that generative models could theoretically describe the true data distribution perfectly. Recent works show for different domains that synthetic data through deep generative modelling can improve downstream models \citep{ PrasannaDas2021ConditionalData, dina2022effect, das2022conditional,  Jain2022SyntheticLearning}, especially for making predictions on small subgroups \citep{bing2022conditional,antoniou2017data}. The latter is important, because small subgroups may correspond to underrepresented minorities. 

\textbf{Challenges.} The challenge of generative data augmentation is that the synthetic data may not be perfect. There is thus a trade-off when using synthetic data for augmentation, between data quantity (i.e. increasing dataset size) versus data quality (i.e. creating potentially unrealistic data). There is some idea of why generative data augmentation helps downstream tasks---see Section \ref{app:augmentation}---but it would be beneficial to gain deeper understanding of \textit{when} synthetic data augmentation aids models and which choice of generative model (hyperparameters) is most suitable.  Ideally, this will result in guidelines that are granular and dataset-dependent, e.g. augment these minority classes, but not the larger class for which there is already enough data. Navigating the quality-quantity trade-off is closely tight in with measuring synthetic data quality, which itself is non-trivial (see Section \ref{sec:challenges}). 

\takeaway{Synthetic data can be used to increase the real dataset size, which has been shown to lead to better downstream models---especially for underrepresented groups in the population.}

\subsection{Domain adaptation} \label{sec:adaptation}
\textbf{Motivation.} In this section, we assume a common scenario: we have some data for the domain of interest $T$(arget), but most data we have comes from other domain(s) $S$(ource) with different distribution(s). The objective of generative model domain adaptation (DA) is to use the source data to create more realistic data for the target domain. The added benefit of using DA at the generative level---versus the downstream model level---is that standard downstream models can be used and compared more easily.

The application of DA for generative modelling is more extensive than it seems at first sight. For example, if there is a large dataset that has some undesired characteristic (e.g. poor minority representation) and a small dataset that has the desired characteristics (e.g. is unbiased), then we can use both datasets to create realistic data with the desired (e.g. \emph{fairness}) properties \emph{without having to define explicitly what these properties are}. Additionally, DA can help overcome \emph{distributional shifts} between training data (e.g. data from country A) and real-world roll-out (country B), by generating more synthetic target data using a small target but large source dataset. Generating synthetic data may also \emph{save resources}, if data is hard or expensive to acquire in the domain of interest. 

\textbf{Challenges.} Most of the current literature focuses on the image setting (see Appendix \ref{app:domain_adaptation}). One of the remaining technical challenges is how to extend generative model domain adaptation to non-image settings, where we have little data from the domain of interest. This is important, because tabular data is predominant in many high-stake applications, e.g. credit scoring and medical forecasting \cite{borisov2021deep,Shwartz-Ziv2022TabularNeed}, and used most often by data scientists\cite{Kaggle20172017Survey}. Another challenge is better understanding of when and what we can transfer. This could allow more specialised transfer of knowledge and structure between domains. For example, transfer feature relationship $A\rightarrow B$ to the new domain, but not $B\rightarrow C$. Additionally, it could facilitate choosing between different available datasets---e.g. when generating synthetic medical data for hospital A, use data from hospitals B and C, but not hospital D.

\takeaway{Generating synthetic data using domain adaptation allows generating synthetic data for settings for which we have little data, which can lead to more accurate and cost-efficient ML.}
\subsection{Data-driven simulations} \label{sec:simulation}
\textbf{Motivation.} The previous two sections assume access to (some) data from the target distribution and aim to generate more data, but in some cases we may have no target data at all. Why are we interested in settings for which we have no data? One of the main applications is model testing. The world is constantly evolving, hence any trained model may operate in a setting with a different data distribution as the training set---typically leading to overestimated real-world performance \citep{patel2008investigating, recht2019imagenet, pianykh2020continuous,quinonero2008dataset}. Ideally, when we have trained an ML model (e.g. a classifier), we want to understand, test, and document how it will behave in unseen scenarios---e.g. in future scenarios or on different populations \citep{gebru2021datasheets}.\footnote{This is also reminiscent of more mature industries, for instance, car manufactures make use of wind tunnels and crash tests to meet regulation standards, whilst electronic component data sheets outline conditions where reliable operation is guaranteed. However, we cannot always go into the real-world to find these datasets.}

Generative models can help to simulate these unseen scenarios based on (e.g. historic) data. Specifically, we can use real data to learn some characteristics in the data, e.g. relationships between selected features, but use a priori knowledge, assumptions, and constraints to modify the overall distribution. For example, we may decide to model the source data using a causal generative model and use do-operations to model interventions \cite{Pearl2009Causality}. In the image domain, another option is to use non-data-driven synthetic data. For example, use CGI-generated images of traffic accidents and add realism through style transfer \citep{wang2019LearningWild}. In contrast to manually searching for other datasets in the ``wild'' \cite{koh2021wilds,hendrycks2018benchmarking}, simulating data through a deep generative process may provide \emph{realistic} data that is relatively \emph{low-effort}, \emph{cost-efficient}, and highly \emph{customizable}. Long term, we believe synthetic data may provide practitioners with a standardized process to train, test ML, and select model(s) under a variety of operating conditions---aiding ML performance, reliability and public acceptance.

\textbf{Challenges.} There are two main challenges for simulated data. The first is reliability of synthetic data itself. Model testing or training on synthetic data is only as accurate and reliable as the synthetic data we have generated, and understanding and quantifying the synthetic data quality is hard---we will discuss this in Section \ref{sec:challenges}. Nonetheless, synthetic data can still quantify when a model \emph{might} fail, which could help model developers decide better which real (expensive) data to acquire to test this hypothesis.

The second challenge is defining meaningful shifts and constraints for developing standardized testing procedures. This requires understanding distributional shifts better on a case-by-case basis. For example, if we want to predict how a trained model will do in the future, we need to understand how the population and feature relationships may change over time. Ideally, the error in these assumptions is propagated to the data, and in turn to the final analyses. 

\takeaway{Synthetic data can be used to simulate domains for which there is no data. In the future this may form a standardized process for characterizing ML model behaviour, leading to higher ML reliability, trustworthiness and acceptance. Propagating uncertainty in the simulation process to the the validity of downstream results is a key challenge.}

\subsection{Bias and Fairness} \label{sec:fairness}
\textbf{Motivation.} Deployed machine learning models have been shown to reflect the bias of the data on which they are trained  \citep{tashea_17AD,dastin,Lu2018contextual_bias,manela2021stereotype,kadambi2021achieving}. This not only affects the discriminated individuals, but also damages society's trust in machine learning as a whole. 
Fairness has thus become an important topic for study in ML. This is especially true for large lanuage models, that are trained on vast amounts of real-world data that inevitably contain some biases, are already being used in practice, and which are so large that bias is harder to detect.

Data users creating models based on publicly available data may be unaware they are inadvertently including bias in their model, or insufficiently knowledgeable to remove it. Generative models have the potential to play an important role in this area, through generating synthetic data based on unfair real data that does not reflect the same biases.\footnote{In its core, this is a form of simulation (Section \ref{sec:simulation}), where we simulate a fair world based on a particular definition of fairness. We discuss it separately, because defining fairness is an important and much-studied topic by itself and comes with a number of unique caveats.} By generating one (or multiple\footnote{A data publisher could publish different synthetic datasets with different fairness constraints.}) fair dataset(s), a data publisher can guarantee {\it any} downstream model satisfies the desired fairness requirements.

\textbf{Defining fairness.} Fairness itself is a broad topic. One face of fairness is representation---some groups may not be represented properly in the data. This is a form of imbalance, for which augmentation (Section \ref{sec:augmentation}) of underrepresented groups can help. Here we focus on a different form of fairness: algorithmic fairness. In its most basic form this considers how some algorithm $A$'s outcome (e.g. whether someone should get parole) depends on some attribute $S$ (e.g. ethnicity). Algorithmic fairness can be further subdivided into different notions of fairness, e.g. direct discrimination and indirect discrimination. \footnote{We refer interested readers to existing surveys for an overview (e.g. \citealp{Mehrabi2021ALearning}).} In \citep{zemel2013learning, xu2018fairgan,xu2019achieving,vanBreugel2021DECAF:Networks}, authors show that by carefully constraining a generative model---with constraints given by the fairness requirement---it is possible to generate fair synthetic data based on unfair real data.

\textbf{Challenges.} Some challenges remain. First, a synthetic dataset that seems fair does not necessarily guarantee that a model trained on this data gives fair downstream predictions---we give an example and explanation in Appendix \ref{app:fairness}. This can be solved \citep{vanBreugel2021DECAF:Networks}, but requires knowledge of the downstream model's deployment setting, which the data publisher does not always have. 

Second, debiasing itself can induce significant data utility loss. This is to be expected: a good but unfair predictor chooses to intentionally use information that it should not, so not allowing this can decrease its performance. As a result, preventing bias entirely may not be desirable. Instead one may be more inclined to use a threshold. For example, the US Supreme Court's 80\% rule \citep{Alessandra1988WhenTrust} essentially states that a prediction has disparate impact if for disadvantaged group $A=1$ and positive outcome $\hat{Y}=1$, $\frac{P(\hat{Y}=1 \vert A=1)}{P(\hat{Y}=1 \vert A=0)}<0.8$ 
\citep{feldman2015certifying}. 
A non-binary vision on fairness also allows balancing multiple fairness notions. This is often necessary, because usually different fairness notions cannot be achieved at the same time, and human subjects do not always find the same notion the most fair \citep{Saxena2020HowAllocations}. 

Third, another challenge is defining bias when the sensitive attribute is not explicitly included. For example, in computer vision, there is not a single pixel that reflects ``asian''. In these domains, human annotations and/or secondary ML systems can be developed to quantify and in turn mitigate bias.\footnote{Ideally, however, fairness is already taken into account at the acquisition stage---acquiring sensitive attributes (e.g. ethnicity and gender) directly allows more accurate understanding and mitigation of bias.}

\takeaway{One can generate \emph{fair} synthetic data based on \emph{unfair} real data, which in turn can be used for training fair ML models.}

\subsection{User-prompted synthetic data} \label{sec:prompted}
\textbf{Motivation.}
Recently, data-driven synthetic data has grasped the imagination of the world through text and image user-prompted generative models---most notably OpenAI's DALL-E 2 \cite{Ramesh2022HierarchicalLatents} and ChatGPT models. Applications of this type of synthetic data are already far-stretching, including query-answering, editing, coding, and artistic aids. These models are also have the potential to reduce existing inequalities, e.g. in healthcare and education \cite{Gates2023TheBegun}.

But should we actually be thinking of these methods as synthetic data generators? Ostensibly, there are large differences with previous use cases. For user-prompted synthetic data the aim is not to replace real datasets with synthetic datasets, but rather assisting humans through providing just one or a few samples that satisfy the prompt---in contrast to providing a whole dataset as output. This also translates into different requirements and metrics for the model output. For example, it may not be desirable that running ChatGPT multiple times would give very different responses, while diversity in synthetic datasets is often desirable \cite{Alaa2022HowModels}.\footnote{From a more mathematical perspective: whereas other synthetic data use cases require approximating a distribution closely, user-prompted synthetic data may be satisfactory if it only achieves a point estimate of the conditional (e.g. the mode of $p(output\vert prompt)$.} On the other hand, user-prompted synthetic data approaches generate data conditionally on (usually text-based) prompts, using deep generative models trained on large quantities of data---thereby satisfying our definition of data-driven synthetic data. More importantly however, we choose to include these methods in this perspective, because they are plagued by some of the same challenges, e.g. the trustworthiness of generated content. 
 
\textbf{Challenges.}
The influence of user-prompted synthetic data has already reached far beyond academic circles. It poses unique challenges---concerning copyrights \cite{Vincent2023GettyVerge}, academic writing \cite{Thorp2023ChatGPTAuthor}, and the future of education \cite{Susnjak2022ChatGPT:Integrity}---but also exemplifies more general AI questions---e.g. transparency, trustworthiness, and accountability \cite{vanDis2023ChatGPT:Research}. Many of these issues can be traced back to the core difficulties of synthetic data generation: quantifying the quality (including factual correctness) and authenticity. These methods are already being deployed, which warrants urgency in solving some of the key synthetic data issues. As an ML community we cannot be compliant and hope someone will fix problems when they arrive; one big-scale failure can have dire consequences, not least for the public perception of AI---especially since models like ChatGPT are the first AIs people actively engage with. This is also the silver lining; widespread interest and high anticipated commercial value provide resources to solve some of these issues. By contextualising methods like ChatGPT as data-driven synthetic data methods, efforts can be centralized into solving some of these issues.

\takeaway{User-prompted synthetic data is a
fast growing field with extensive applications. Its already-widespread use implores urgent solutions to some of the key, more general synthetic data challenges.}

\section{General challenges and opportunities} \label{sec:challenges}
\begin{figure*}
    \centering
    \includegraphics[width=\textwidth,angle=0]{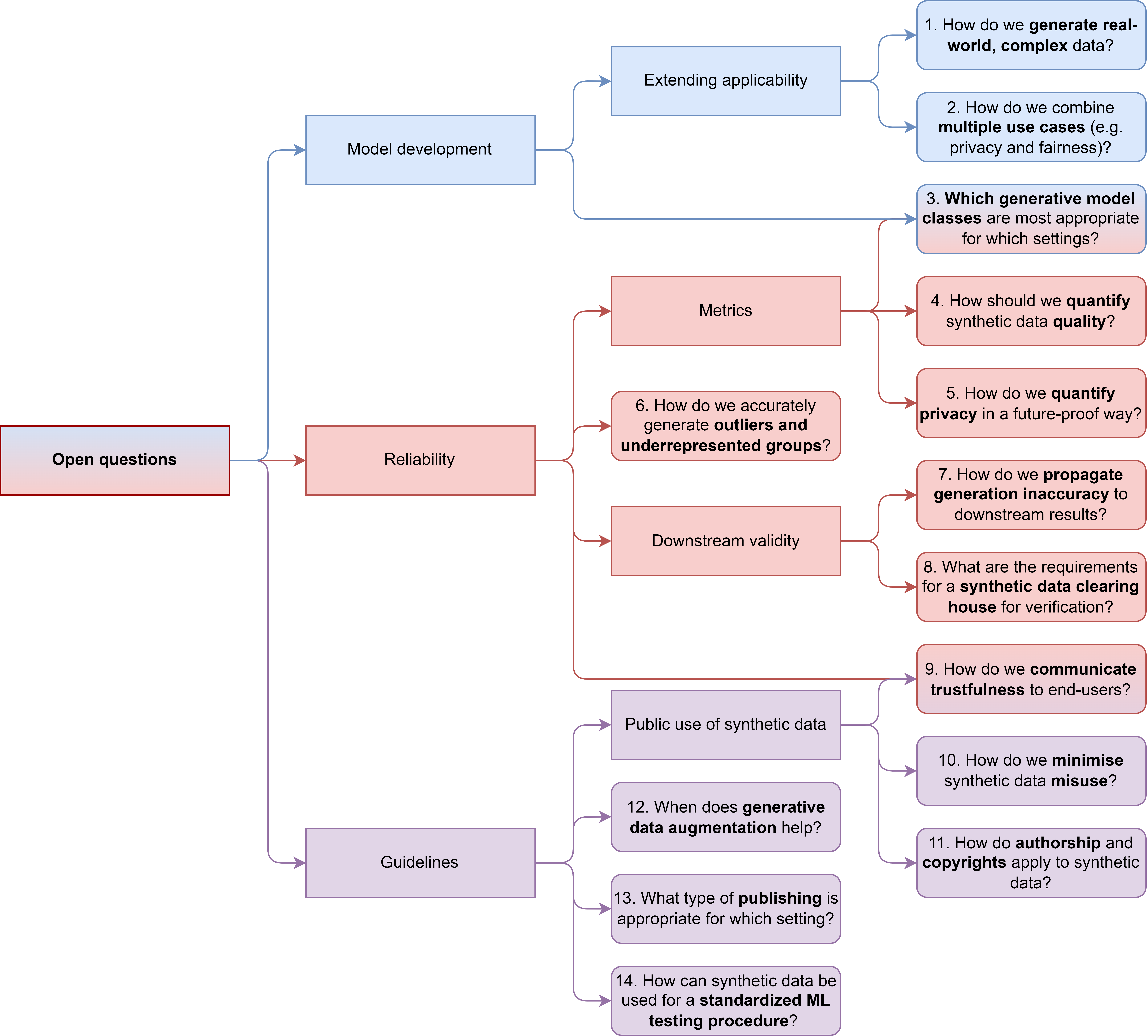}
    \caption{Selection of open questions in the synthetic data realm. Numbering does not indicate perceived importance.}
    \label{fig:challenges}
\end{figure*}

Let us now discuss the general challenges for widespread synthetic data use. In Figure \ref{fig:challenges} we include 14 major open questions, and we refer to these in text like this: \oq{{\tiny\#}}.

\textbf{Extending applicability.} \oq{1} Some settings are significantly underexplored in the generative modelling literature, including high-dimensional tabular data, multimodal data (e.g. medical records consisting of text and tabular), relational databases, and \oq{2} combining use cases (e.g. fairness \emph{and} privacy). Extending applicability increases the potential of synthetic data for both scientific and industrial use.

\textbf{Metrics.} Trust in synthetic data requires measurability of desired properties---quality, privacy, etc. Not only is it hard to measure some of these properties, even defining them is a challenge. Current metrics for \oq{4} quality and \oq{5} privacy are insufficient, see Table \ref{tab:metrics} in Appendix \ref{app:metrics} for an overview. We envision five properties for better synthetic data metrics. First, metrics should provide insight into different aspects of synthetic data (e.g. fidelity, diversity, and authenticity as in \citep{Alaa2022HowModels}). Secondly, they should be interpretable to a downstream user. Interpretability allows them to decide whether to trust the downstream analysis, and also helps choose a good trade-off between competing metrics (e.g. privacy-utility or fairness-utility). Third, the ideal metric should allow for granular evaluation; instead of just knowing how ``good'' a synthetic dataset is overall, we would like to know whether some groups are misrepresented, more vulnerable to privacy attacks, etc. Fourth, it should be reliable and robust, enabling trustworthy benchmarking and providing consistent results across studies and implementations. Fifth, it should be data-efficient and robust to varying amounts of data available. For example, many metrics (e.g. \citep{Yoon2020AnonymizationADS-GAN}) require that real and synthetic datasets are the same size, but real data is usually limited whereas synthetic data can be generated freely. As long as no set of metrics exists that satisfies all these desiderata, the use of synthetic data will be severely limited. 

\textbf{Which generative model to use?} A plethora of generative models are being developed, but there is little insight into \oq{3} which model should be used when. Evidently, having better metrics would aid this process by facilitating generative model selection---potentially trading off different metrics depending on the downstream use case. This can be aided further through model standardization and more unified generative modelling libraries  (e.g. see \cite{Qian2023Synthcity:Modalities}), including more benchmarking datasets for different generative modelling settings.

\textbf{Outliers and underrepresented regions.} Low-density regions are more difficult to model, because by definition there are fewer samples to learn from. This is especially true when privacy constraints are used, since these methods limit the influence of any single point. This poses a problem for synthetic data, since low-density regions are more likely to correspond to outliers---which can be of primary interest to scientists---and minority groups---incorrect modelling of which could lead to discrimination. On the positive side, Section \ref{sec:augmentation} explored how generative models have been shown to in fact \textit{improve} the model performance in low-density regions, compared to using the real data only. \oq{6} Understanding the limitations of generative modelling in data-scarce regions remains an underexplored challenge. 

\textbf{Understanding influence of synthetic data on downstream task.}  Going further than metrics, \oq{7} we want to understand \textit{how} the synthetic data step affects our results. For example, errors and uncertainty in the generative process should be propagated to downstream uncertainty quantification and significance tests.

\textbf{Data clearing house for verification protocols.} Until the above-mentioned reliability challenges are resolved, acceptance of results derived from synthetic data will be low and hence require some form of verification. Verifying results on real data is not always possible or desirable however, because data users may not have access to real data, and data publishers do not have the capacity to verify any end-user's result. \oq{8} One possibility is to create a protected data entity, which \textit{does} have access to real data and can automatically run tests. We envision this as an API that can answer queries, yet is carefully restricted from leaking too much information of the real data---e.g. users need to be verified, cannot test too many hypotheses, and the API's output is partly hidden or noisy. 

\textbf{Publishing and access.} At last, \oq{13} how should we publish synthetic data. There are three options: (i) publishing the generative model, (ii) publishing only the data, and (iii) API access to generate based on some user input (e.g. ChatGPT). Each has advantages and its limitations (see Appendix \ref{app:publishing}) and data publishers may choose different forms of access dependent on the context.

\section{Conclusion}
Synthetic data offers a broad range of advantages over real data; from privacy to user-prompted generation aids, and from better (more accurate, robust, and fair) downstream models to cost-efficient and more versatile use of real data. We believe long-term, standardized procedures for generating synthetic data will become key components to model training, testing, and benchmarking. For synthetic data to really change the ML world, however, key challenges remain. We urge the community to shift resources; focusing less on creating stronger generative models (e.g. that create more realistic pictures), towards seeking to understand the quality of synthetic data and its effect on the validity of downstream results or predictions. With the world's eye caught by generative models, fundamental understanding in trustworthiness of synthetic data is urgently required.

\bibliographystyle{plainnat}
\bibliography{mendeley, references_non_mendeley}

\clearpage
\appendix
\section{Metrics} \label{app:metrics}
There exist a range of metrics for measuring synthetic data quality and privacy. In Table \ref{tab:metrics} we include a non-exhaustive list.

\begin{table*}[hbt]
    \centering
    \caption{Metrics for synthetic data. }
    \begin{tabularx}{\textwidth}{c|X} \toprule
        Metric & Description \\ \midrule
         \multicolumn{2}{c}{Quality} \\ \midrule
         WD \cite{Arjovsky2019InvariantMinimization}& Wasserstein Distance aims to approximate the shift from the real to synthetic data distribution, but it is not intuitive for end-users and requires training a model for computation, which can be unstable.\\ \hline
         FID \cite{Heusel2017GANsEquilibrium} & 
         FID scores correspond fairly well with human quality assessments, which have made them the most popular metric in image applications. Its reliance on InceptionV3 limits its use to other data modalities, or to imaging settings (e.g. medical images) which do not have similar classes as ImageNet on which InceptionV3 was trained.\\ \hline
         NND \cite{Arora2017GeneralizationGANs} & Neural net distance trains critic to distinguish between real and fake samples, and uses its success as a metric. This is flexible, intuitive, and extendable to different modalities, but is highly dependent on the critic's implementation.\\ \hline
         \makecell[ct]{Precision and\\ Recall \cite{Sajjadi2018AssessingRecall,Kynkaanniemi2019ImprovedModels}}&  Implicitly assumes Euclidean distance for computing quality, which scales badly to higher dimensions and may not work well for all data modalities.\\ \hline
         \makecell[ct]{Density and\\ Coverage \cite{Naeem2020ReliableModels}}& Improves upon the previous work through better theoretical guarantees, but has the same limitations.\\ \hline
         \makecell[ct]{$\alpha$-Precision and\\$\beta$-Recall \cite{Alaa2022HowModels}}& More intuitive interpretation of fidelity and diversity, is compatible with different modalities, and does not assume a Euclidean metric. However, requires training a OneClass model for embedding the data into a space where the metric is computed, which may be unstable, is implementation dependent, and may not work for all types of data (e.g. multimodal data).\\ \hline
         \makecell[ct]{Downstream model\\ performance}& Can be used in conjunction with any downstream model and metric. This is an advantage, allowing flexibility in testing different modalities, but requires that the data publisher knows the downstream user's task and model, which is usually not the case, and requires time and expertise on from the data publisher.\\ \midrule
        \multicolumn{2}{c}{Privacy} \\ \midrule $\epsilon$-Identifiability \cite{Yoon2020AnonymizationADS-GAN}& Heuristic to compute generalization based on distance of real data to synthetic vs other real points.\\ \hline
         k-anonymity \cite{Sweeney2002K-anonymity:Privacy}& Does not work for continuous features and focuses on anonymised data where linkage attacks are possible---this is usually not an issue for synthetic data.\\ \hline
         $\ell$-diversity \cite{Machanavajjhala2006-Diversity:K-anonymity}& Aims to extend k-anonymity, through requiring more diversity in sensitive variable for quasi-identifier (QID) equivalence groups. This means an attacker cannot with absolute certainty determine any patients' sensitive attribute. Focuses on anonymised data.\\ \hline
         t-closeness \cite{Ninghui2007T-Closeness:-diversity}& Aims to extend $\ell$-diversity through not just requiring more diversity, but more entropy. This means an attacker cannot with high certainty determine any patients' sensitive attribute. Non-intuitive. Focuses on anonymised data. \\ \hline
         ($\alpha$, k)-anonymity \cite{Wong2009Publishing}& Argues to be more intuitive than $t$-closeness. Focuses on anonymised data\\ \hline
         \makecell[ct]{Privacy attacker\\ performance \\ \cite{Hayes2019LOGAN:Models,Hilprecht2019MonteModels, Chen2019GAN-Leaks:Models, Liu2019PerformingModels, Hu2021MembershipRegions, vanBreugel2023domias}}& Any architecture or task can be used, e.g. membership, difference, reconstruction and linkage attacks, though at the time of writing all generative MIA works focus on membership inference. Does not give guarantees and is dependent on the choice of attacker architecture and access.\\ \hline
         \makecell[ct]{$(\epsilon, \delta)$-Differential\\ privacy \cite{Dwork2014ThePrivacy}} & Theoretical guarantees on the influence of data on the final model. Is not available as a metric and is often too strong (privacy is only achieved with high data utility loss). It is also hard to actually set the privacy budget $\epsilon$ as this is highly non-intuitive \cite{Lee2011HowPrivacy}\\ \bottomrule
    \end{tabularx}
    \label{tab:metrics}
\end{table*}

\section{Privacy: existing work}
\label{app:privacy}
\textbf{Generating privacy data.} 
Differential Privacy (DP) \citep{Dwork2006OurGeneration} is by far the most used mechanism for generating private data \citep{Zhang2017Privbayes:Networks, Jordon2019PATE-GAN:Guarantees, Ho2021DP-GAN:Nets, Torkzadehmahani2020DP-CGAN:Generation}. Essentially, DP limits how much any real sample can influence the final dataset. Formally, it is defined as follows:
\begin{definition}
    \citep{Dwork2006OurGeneration} A randomized algorithm $M:\mathcal{D}\rightarrow \mathcal{R}$ with domain $\mathcal{D}$ and range $\mathcal{R}$ satisfies $(\epsilon,\delta)$-differential privacy if for any adjacent inputs $d,d' \in \mathcal{D}$ and for any $S\subseteq \mathcal{A}$ it holds that:
    \begin{equation*}
        \Pr[\mathcal{M}(d)\in S]\leq e^\epsilon \Pr[\mathcal{M}(d')\in S) +\delta.
    \end{equation*}
\end{definition}
The main advantage of using DP is that it gives theoretical guarantees on the privacy of the synthetic data, and by extent any downstream analysis, but the clear disadvantage of DP is that the definition is non-intuitive and parameter $\epsilon$ is hard to choose. Instead, some authors opt for more heuristic privacy notions, e.g. \citet{Yoon2020AnonymizationADS-GAN} consider how close the synthetic data is to the real data in Euclidean space. Another branch of research takes an adversarial approach and studies privacy attacks themselves \cite{Hayes2019LOGAN:Models,Hilprecht2019MonteModels, Chen2019GAN-Leaks:Models, Liu2019PerformingModels, Hu2021MembershipRegions, vanBreugel2023domias}; by understanding better methods that actively aim to retrieve information about the training data, the hope is we may get practical insight into privacy itself. These empirical studies seem to show the privacy risk of a properly trained generative model without DP constraints, where only the synthetic data (and not the model) is published---is usually quite low overall. Nonetheless, higher risk is possible for outlier groups \citep{Hu2021MembershipRegions, vanBreugel2023domias} and when (in the future) more information becomes available  \citep{vanBreugel2023domias}. Overall, guaranteeing privacy is hard, which brings us to the challenges.

\textbf{Attacking models} Measuring privacy is hard. For example, though a generative model \emph{itself} could be DP, there is no method to measure this post-hoc on the data. Surprisingly, this research is still fairly underdeveloped. The most studied attack is also the simplest one, Membership Inference \citep{Shokri2017MembershipModels}: can an attacker determine whether some sample $x^*$ they possess, was used for generating the synthetic data? Most of the existing attacker models \citep{Hayes2019LOGAN:Models, Hilprecht2019MonteModels, Chen2019GAN-Leaks:Models, Liu2019PerformingModels}, however, are no better than random guessing unless the attacker has some white-box information---i.e. knowledge of the internals of the generator. This can often be avoided in practice, since a data publisher usually only publishes the data, not the generative model. Recent works, however, have shown that in some scenarios attackers can be successful. \cite{Hu2021MembershipRegions} have shown that high-precision attacks are possible against low-density regions. This is problematic, since these regions are more likely to correspond to minority groups. \cite{vanBreugel2023domias} have similar findings, and show that attackers can initiate significantly more successful attacks when they have access to an independent dataset.

\section{Data augmentation: extended discussion}
\label{app:augmentation}
\textbf{Tabular data augmentation.}  Traditional methods in this domain oversample data, add Gaussian noise \cite{Bishop1995NeuralRecognition}, or interpolate \citep{Chawla2002SMOTE:Technique,Han2005Borderline-SMOTE:Learning,He2008ADASYN:Learning}, which can effectively balance imbalanced datasets and limit overfitting of the downstream task. However, in almost all cases these augmentation approaches lead to some data utility loss, since the outputted distribution is noisier than the true data distribution. The advantage of generative data augmentation over traditional tabular data augmentation  methods \citep{Chawla2002SMOTE:Technique,Han2005Borderline-SMOTE:Learning,He2008ADASYN:Learning} is that generative models could theoretically describe the true data distribution perfectly when there is enough data to learn from---cf. naive interpolation. 

\textbf{Why it works.}
Though promising, we should be somewhat skeptical of generate data augmentation. In contrast to typical data augmentation used in domains like imaging, we do not add any a priori knowledge. Training generative models usually requires large amounts of data and in any case may not describe the real data perfectly, so it is unclear why using a synthetic dataset $D_S$ generated by a generator trained on some dataset $D_R$, could be better than just using the data $D_R$ directly. Nonetheless, this is exactly what the aforementioned studies show. \citet{antoniou2017data} provide a possible explanation: the higher-dimensional distribution may live on a lower-dimensional manifold, which the generative model is able to learn. As a result, the generative model may approximate the real data very well, even in regions with apparently low density (when viewed in the high-dimensional space). Another constituting factor may be that the advantage of having additional data for the downstream model (making it less likely to overfit, effectively regularizing it), outweighs the potential disadvantage of possible noise that the generative modelling step induces. This is similar to the popular technique of adding noise to a neural networks input,\footnote{This is technically equivalent to synthetic data augmentation using a gaussian kernel density estimator for approximating the distribution (i.e. approximating the distribution by putting a Gaussian on top of each real data point).} often resulting in better regularized and more robust models.

\section{Domain adaptation: existing work} \label{app:domain_adaptation}
There has been considerable attention for domain adaptation (DA) in generative modelling. One way to generate data across domains is to translate samples from one domain to the other. Even though this is inherently an ill-posed problem \citep{Lindvall2002LecturesMethod,Liu2017UnsupervisedNetworks, Liu2019Few-ShotTranslation}, this has become a popular area of research with good results in the image setting. CoGAN \citep{Liu2016CoupledNetworks}, discoGAN \citep{Kim2017LearningNetworks} and cycleGAN \citep{Zhu2017UnpairedNetworks} have popularised GAN image-to-image translation. This was followed by many improvements, e.g. allowing for more than two domains \citep{Liu2017UnsupervisedNetworks, Choi2017GeneratingNetworks,Choi2019StarGANDomains,Anoosheh2017ComboGAN:Translation}. The problem with these works is the large amounts of data required to train each domain's generator. StyleGAN \citep{Karras2018ANetworks} and follow-up works \citep{Karras2019AnalyzingStyleGAN,Karras2020TrainingData} solve this by distinguishing the style and content of an image, and single-shot transferring the style from one image to another---e.g. conditionally creating the painting equivalent of some photo in the style of Van Gogh's Starry Night. Unfortunately, domain adaptation is much harder for non-image modalities (e.g. tabular). Style transfer is often not applicable, because there may be no clear distinction between style and content. RadialGAN \citep{Yoon2018RadialGAN:Networks} extends \citep{Choi2017GeneratingNetworks} to tabular medical data, but it is severely limited by requiring a large quantity of data in the target domain. More research is required that considers small domains, in particular small target domains.

\section{Fairness: extended discussion} \label{app:fairness}

\textbf{Fairness notions.}
Algorithmic fairness can be further subdivided into different notions of fairness, e.g. direct discrimination and indirect discrimination. We will not go into different definitions, instead referring readers to existing surveys (e.g. see \citealp{Mehrabi2021ALearning}) and highlighting that choosing a definition requires careful public debate on a case-to-case basis.\footnote{Let us give the example of affirmative action \citep{Holzer2000AssessingAction}. Applicants A and B have roughly the same education and apply to the same prestigious school. Should the school favour one over the other? Without any other information, no. But what if applicant B has an ethnicity that is very poorly represented in the school? Whether this legitimizes selecting applicant B over A is both ethically and legally complex \citep{Bent2019IsLegal}.} Given some notion, however, ML researchers and practitioners can aim to find and mitigate unfairness.

\textbf{Data that looks fair, may not be fair.}
Generating fair synthetic data is not simple: a synthetic dataset that seems fair does not necessarily guarantee fair downstream predictions. Let us give an example.

\begin{example}
Let us assume there is a dataset with 4 binary variables with graphical model $A\rightarrow B \rightarrow Y \leftarrow C$, with $A$ a sensitive attribute and $Y$ the target we want to predict using some prediction model $g$. Let us suppose we want to prevent indirect discrimination by downstream users. We decide to create a new synthetic dataset, where we simply copy the real data, except for variable $A$ which we choose randomly. Evidently, $A$ is uncorrelated with all the features and the target in this synthetic dataset, so a downstream user will think it is ``fair''. The downstream user trains predictive model $g$ and finds that feature $B$ is a strong indicator of $D$, and that predictions are uncorrelated with $A$. They deploy the model confidently, but find that the model contains high amounts of indirect discrimination against $A$ on real data.
\end{example}

The problem, in this example, is that synthetic data \emph{looking fair} by itself, is not the same as models trained on this data \emph{being fair on a different (e.g. real) data distribution}.  \citet{vanBreugel2021DECAF:Networks} introduce a causal generative model (DECAF) that guarantees that the optimal downstream predictor is fair.

\section{Publishing data}
\label{app:publishing}
There are three options on how to provide access to synthetic data: (i) publishing the generative model, (ii) publishing only the data, and (iii) API access to generate based on some user input (e.g. ChatGPT). Each has advantages and its limitations. (i) Publishing the generative model allows a user to generate as much data as they like, fine-tune models on their own data, and directly choose the input to the generative model---e.g. generate data for different fairness notions. This is also relevant for interactive settings like reinforcement learning, where access to a synthetic environment (versus only offline data) would be highly beneficial to the end-user. In this case, it may simply not be possible for a data publisher to generate and store different copies of synthetic data for each possible downstream application or policy. 
(ii) On the other hand, publishing synthetic data directly will be easier for many users and settings, as it does not require expertise on how run generative models correctly. The second option is also better when there are privacy concerns; access to the internals of a generative model can disclose significantly more information about the training data \citep{Chen2019GAN-Leaks:Models}. (iii) API access could be the best of both worlds---enabling end-users to provide some form of input in an accessible way, with better privacy guarantees than publishing the model itself. On the other hand, creating and hosting an API of the generative model requires technical expertise and financial means on the side of the generative model owner. 

\end{document}